\documentclass[sigconf, nonacm]{acmart} 

\setcopyright{none}
\acmDOI{}
\acmISBN{}
\acmConference{}
\makeatletter
\def\@copyrightspace{} 
\makeatother

\usepackage{fancyhdr}
\fancypagestyle{firstpage}{%
  \fancyhf{}
  \fancyfoot[C]{\textit{Preprint submitted to arXiv}}
}
\usepackage{graphicx} 
\usepackage{enumitem}

\AtBeginDocument{%
  }

\begin{document}

\title{FedFitTech: A Baseline in Federated Learning for Fitness Tracking}

\author{Zeyneddin Oz}
\orcid{0000-0002-4216-9854}
\authornotemark[1]
\affiliation{%
  \institution{Ubiquitous Computing, University of Siegen}
  \city{Siegen}
  \state{NRW}
  \country{Germany}
}
\email{zeyneddin.oez@uni-siegen.de}

\author{Shreyas Korde}
\affiliation{%
  \institution{Ubiquitous Computing, University of Siegen}
  \city{Siegen}
  \state{NRW}
  \country{Germany}
}
\email{shreyas.korde@student.uni-siegen.de}

\author{Marius Bock}
\orcid{0000-0001-7401-928X}
\affiliation{%
  \institution{Ubiquitous Computing, University of Siegen}
  \city{Siegen}
  \state{NRW}
  \country{Germany}
}
\email{marius.bock@uni-siegen.de}

\author{Kristof Van Laerhoven}
\orcid{0000-0001-5296-5347}
\affiliation{%
  \institution{Ubiquitous Computing, University of Siegen}
  \city{Siegen}
  \state{NRW}
  \country{Germany}
}
\email{kvl@eti.uni-siegen.de}

\renewcommand{\shortauthors}{Oz et al.}

\begin{abstract}
The rapid evolution of sensors and resource-efficient machine learning models has spurred the widespread adoption of wearable fitness tracking devices. Equipped with inertial sensors, such devices can continuously capture physical movements for fitness technology (FitTech), enabling applications from sports optimization to preventive healthcare. Traditional Centralized Learning approaches to detect fitness activities struggle with data privacy concerns, regulatory restrictions, and communication inefficiencies. In contrast, Federated Learning enables a decentralized model training by communicating model updates rather than potentially private wearable sensor data. Applying Federated Learning to FitTech presents unique challenges, such as data imbalance, lack of labeled data, heterogeneous user activities, and trade-offs between personalization and generalization. To simplify research on FitTech in Federated Learning, we present the FedFitTech baseline, under the Flower framework, which is publicly available and widely used by both industry and academic researchers. Additionally, to illustrate its usage, this paper presents a case study that implements a system based on the FedFitTech baseline, incorporating a client-side early stopping strategy and comparing the results. For instance, this system allows wearable devices to optimize the trade-off between capturing common fitness activities and preserving individuals' nuances, thereby enhancing both the scalability and efficiency of privacy-aware fitness tracking applications. The results show that this reduces the overall redundant communications by 13\%, while maintaining the overall recognition performance at a negligible recognition cost by 1\%. Thus, the FedFitTech baseline creates a foundation for a wide range of new research and development opportunities in FitTech, and it is available as open source at: \url{https://github.com/shreyaskorde16/FedFitTech}{}
\end{abstract}





\maketitle

\section{Introduction}

With recent advances in sensor technologies and resource-efficient machine learning models, wearable fitness trackers in various forms, such as wrist watches, glasses, earbuds, and rings, have gained widespread use. These devices generate a vast amount of valuable sensor data for Human Activity Recognition (HAR) applications, in which body movements can be analyzed in detail with minimal intrusion. For example, inertial sensors can continuously monitor motion and gestures at specific body locations, thus providing a detailed representation of the user's motion patterns. This makes such devices particularly useful for an extensive range of applications in fitness technology (FitTech), including medical assistance and the optimization of complex work processes \cite{bulling2014tutorial}.

FitTech is a crossover term that combines fitness and technology, encompassing innovations that enhance fitness, wellness, and health experiences. Using wearable devices, mobile apps, and virtual personal training platforms, FitTech can monitor physical activity, provide real-time feedback to users, and tailor exercise programs to their needs. Through machine learning, FitTech promises to transform the fitness industry by making workouts more personalized, efficient, and data-driven, helping users achieve better training results.

Current research tends to focus largely on Centralized Learning of fitness activities, which can be used to develop powerful machine learning models by being able to process large datasets and leverage substantial computational power. However, Centralized Learning requires sensor data to be shared and stored in a central location, which can deter data sharing due to regulatory restrictions (e.g., the European General Data Protection Regulation (GDPR) \cite{voigt2017eu} and the California Consumer Privacy Act (CCPA) \cite{de2018guide}), making users reluctant to grant access for training. Moreover, Centralized Learning struggles to meet the needs for scalability and communication efficiency.

\begin{figure}[t]
    \centering
    \includegraphics[width=\linewidth]{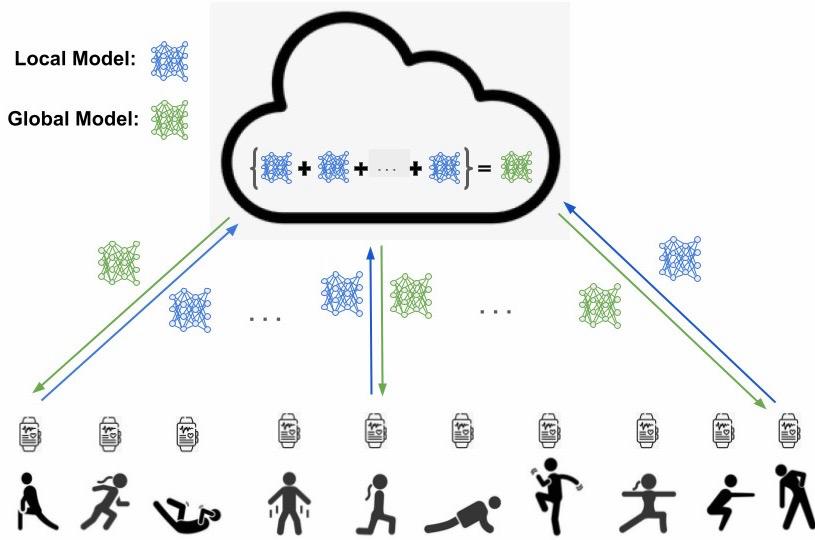}
    \caption{In FedFitTech, wearable fitness devices locally train a global classification model for fitness activities (green model) shared by a server. Tuned local models (blue models) are shared with the server for aggregation to update a global model across all participants. Then, this global model is redistributed for further training rounds, which leads to a private system while profiting from outer patterns.}  
  \label{fig:fedfittech}
\end{figure}

To address these challenges, Federated Learning \cite{mcmahan2017communication} has emerged as an alternative to Centralized Learning, offering a model communication mechanism instead of raw data sharing. In Federated Learning, a global model is shared with devices (also known as clients, nodes, or workers) for local training using local data. These tuned local models are then sent back to the server, which aggregates them to update the global model. The updated global model is redistributed to the devices for further training rounds. At the end of the process, all devices obtain a global model trained on all local data, while data privacy is maintained, as sensitive raw data are not transmitted to external sources (see Fig.~\ref{fig:fedfittech}). Additionally, Federated Learning offers benefits such as improved data availability, scalability, fault recovery, and communication efficiency \cite{liu2021learning}. 

Although Federated Learning is widely recognized as a viable approach for training machine learning models on decentralized devices, its potential impact on FitTech remains underexplored. Wearables such as smartwatches typically share similar sensors, particularly MEMS inertial sensors, as well as processing and wireless networking components, making FitTech applications relatively homogeneous compared to other environments. This similarity results in consistent computation and communication capabilities across smartwatches, enabling training of the same global model with synchronized communication.

Additionally, researchers of \cite{mcmahan2017communication} identified communication as the primary challenge in Federated Learning systems. However, unlike other domains, fitness tracking does not require daily communication. Since people do not exercise frequently throughout the day, data is not generated continuously, and wearable devices accumulate small datasets. This allows for less frequent and more convenient communication times, such as during charging or when connected to unmetered Wi-Fi. Moreover, modern processors make the computation of tiny machine learning models more efficient, reducing costs. Furthermore, authors of \cite{ek2022evaluation} compared various Federated Learning algorithms and found that Federated Learning can achieve comparable or even superior performance to Centralized Learning techniques in the HAR domain. Thus, the FitTech environment is naturally well suited to Federated Learning.

On the other hand, each person exhibits unique movement patterns and the performing of the same exercise can vary depending on individual performance. The position of the sensors on a person can affect the recorded motion patterns, even when repeating the same actions \cite{arikumar2022fl}. Factors such as gender, age, weight, height, and exercise frequency further contribute to these unique characteristics \cite{li2021meta}. For instance, one person might jog once a week, while another person might jog every other day, eventually leading to data imbalances. The literature shows that Federated Learning underperforms when device data is imbalanced, as excessive external patterns can reduce performance on local data due to the personal nature of movements \cite{cheng2023protohar}. Moreover, annotated local data is often limited, requiring patterns from other people to accurately detect a person's exercise. Given the advantages of Federated Learning in FitTech, we thus present in this paper FedFitTech, a Federated Learning baseline designed for research in the FitTech domain.

Furthermore, we give an example using the FedFitTech baseline by designing a client-side early stopping implementation and analyzing its benefits. In FitTech, due to the varied motion patterns of the user, the global model should not override individual variations. Thus, early stopping may be helpful when the global model underperforms on local datasets. The attractiveness of implementing client-side early stopping in FedFitTech is that, in the standard Federated Learning setting \cite{mcmahan2017communication}, all devices participate in the training of the global model for a fixed number of communication rounds. This can lead to unnecessary resource consumption for devices that do not need to attend all training rounds. In addition, the FitTech domain requires incorporating only common patterns from others without losing the influence of personal data in the global model. Although Federated Learning provides a private way to benefit from external patterns, users need the global model to be tailored to local data. Therefore, it is essential to design a Federated Learning system that balances personalization and generalization in FitTech. Thus, our case study of the FedFitTech baseline involves an early stopping method to mitigate these challenges.  

Our contributions are threefold:


\begin{itemize}
    
    \item \textbf{Federated Fitness Activity Recognition:} We highlight the domain of fitness activity recognition (or HAR) as interesting and suitable for the field of Federated Learning.  
    \item \textbf{Flower Baseline for Fitness Tracking:} We designed FedFitTech as an easy-to-use baseline built under the Flower framework \cite{beutel2020flower}, a widely adopted and openly available platform that enables the implementation and benchmarking of reproducible experiments. 
    \item \textbf{Use Case Experiment of Early Stopping:} We present an example usage of this baseline, by allowing client-side early stopping to reduce energy consumption and thus improving the overall efficiency of the system.
    
\end{itemize}

\section{Related Work}

Several studies have investigated various implementations of Federated Learning for HAR, addressing different challenges through novel strategies. FL-PMI \cite{arikumar2022fl} proposes a system leveraging unlabeled data by applying a cleaning process with auto-labeling propagation to support smart healthcare applications. Meta‐HAR \cite{li2021meta} employs a model‐agnostic meta-learning approach that extracts adaptive signal embeddings from sensor data and integrates a task‐specific classification network through fine-tuning. PMF \cite{feng2020pmf} adopts a secure data handling approach by combining differential privacy with secure multi-party computation and introduces an adaptive weight adjustment mechanism to ensure fair contributions to a global model. Also, the work by \cite{xiao2021federated} presents a perceptive extraction network system that enhances data representation while employing encryption and decryption to mitigate potential data leakage risks.

To address the challenges of inadequate data and secure data sharing, researchers in \cite{zhou20222d} propose a two-dimensional framework that employs both horizontal and vertical FL paradigms, while another study \cite{zhou2022you} utilizes specialized training algorithms aimed at increasing data diversity and fostering inclusivity for devices with inferior network conditions.

Focusing on model training improvements, several works explore client grouping and dynamic layer sharing. PS-PFL \cite{chai2024profile} introduces a personalized FL model based on profile similarity that aggregates local models through weighted computations. An attention-based clustering mechanism in \cite{bu2024learn} calculates an optimal weighted model combination to encourage similar clients to group together without relying on a single global model. ClusterFL \cite{ouyang2021clusterfl} leverages a cluster indicator matrix and iterative optimization techniques to update client model weights, facilitating cluster-wise straggler dropout and correlation-based client selection. ProtoHAR \cite{cheng2023protohar} integrates a prototype-guided technique that separates representation learning from categorization, thereby achieving uniform activity feature representation and improved convergence. Also, dynamic layer-sharing is investigated by FedCLAR \cite{presotto2022fedclar}, which selectively shares portions of model weights based on a hierarchical clustering, and by FedDL \cite{tu2021feddl}, which merges lower layers of models to capture user-specific similarities while enhancing communication efficiency.

Furthermore, to enhance privacy, the researcher of PDP-FL (Personalized Differential Privacy-based Federated Learning) \cite{shao2024protecting} specifically investigated the fitness tracking application using two stages. First, it allows users to customize their privacy levels, with corresponding noise added to their local model prior to sharing with a server. The second stage involves the server adding further noise to ensure global privacy protection. 

The weaknesses noted across these studies include that FL-PMI \cite{arikumar2022fl} may require additional computational resources, which could impact battery life, and does not explicitly address sensor inaccuracies. Meta‐HAR \cite{li2021meta} may incur higher computational and communication overhead, posing challenges for very large-scale deployments. While promising, PMF \cite{feng2020pmf} requires further empirical validation of its performance under large-scale data scenarios and the integration of scalable privacy-preserving measures. The two-dimensional framework by \cite{zhou20222d} could potentially expose finer-grained activity details under adversarial conditions, and the approach in \cite{zhou2022you} may face challenges with dynamic client participation and scaling to real-world deployments. Additionally, client grouping strategies such as PS-PFL \cite{chai2024profile} incur extra computational overhead on resource-constrained devices, and ClusterFL \cite{ouyang2021clusterfl} might expose user-cluster associations under adversarial conditions while necessitating an additional optimization step. Lastly, ProtoHAR \cite{cheng2023protohar} assumes a stationary data distribution, which might limit its adaptability in continuously evolving environments, FedCLAR \cite{presotto2022fedclar} assumes the availability of ample labeled data — a condition that may not hold universally — and FedDL \cite{tu2021feddl} requires further integration of robust privacy measures along with additional validation of its effectiveness beyond HAR. Besides, while PDP-FL \cite{shao2024protecting} enhances privacy by allowing varied user privacy preferences, this diversity can impact model performance and effectiveness. In other words, differing noise levels, corresponding to individual privacy choices, influence global model aggregation. To add more, PDP-FL simulates the fitness tracking environment using image data, which does not fully capture the complexities of the domain. Fitness data often have temporal dependencies, varying sampling rates, and different types of noise characteristics compared to image pixels.  

While other studies addressed HAR applications, this study especially focuses on the FitTech domain using FL, similar to PDP-FL. However, our FedFitTech baseline utilizes a real fitness tracking dataset and a tiny model, reflecting original FitTech characteristics and providing further exploration of FL potential in this area.

About designing a case study of FedFitTech, the implementation of early stopping in the FL setting has already been explored in FLrce \cite{niu2024flrce} and FLASH \cite{panchal2023flash}. FLrce employs a resource-efficient early stopping strategy tailored for FL. FLASH leverages concept drift detection through client-side early stopping and server-side adaptive optimization. However, their complex mechanisms increase computational overhead and risk premature termination. Also, these methods are primarily designed for image recognition in non-FitTech domains. Our case study especially focuses on the FitTech domain and employs an early stopping approach based on the stability of the validation F1-score over a sliding window \cite{prechelt2002early}. While this method is simple to implement, it ensures that the global model retains valuable patterns learned from all clients, including those with less significant contributions, which is an essential factor in FL for FitTech. Moreover, the case study emphasizes communication efficiency and balanced generalization. Despite a negligible F1-score drop, our case study allows clients to benefit from external patterns and terminate training when global weights diverge from local patterns, preventing redundant computation and communication.

\section{Experimental Setting}

\subsection{Model}
Over the past decade, Deep Learning (DL) methods have become the dominant approach in inertial-based HAR, consistently outperforming traditional Machine Learning algorithms \cite{ordonez2016deep, guan2017ensembles}. For this reason, we used TinyHAR \cite{zhou2022tinyhar}, which is a lightweight DL model specifically designed for HAR. TinyHAR leverages a combination of convolutional layers, a transformer block featuring self-attention and fully connected layers, and a Long Short-Term Memory (LSTM) to extract and model features from sensor data efficiently. Initially, convolutional layers are applied to each sensor channel to capture local patterns. This is followed by a transformer block that facilitates effective cross-channel feature interaction. Next, a fully connected layer fuses the extracted features, which are then processed by an LSTM to capture long-term temporal dependencies. Finally, a self-attention layer dynamically recalibrates the importance of features across time. This architecture achieves competitive performance with a significantly reduced parameter count, making it highly suitable for deployment on resource-constrained devices.

\subsection{Dataset}
We used inertial-based data from the WEAR (Wearable and Egocentric Activity Recognition) (\url{https://mariusbock.github.io/wear/}, \cite{bock2024wear}) dataset, which consists of labeled activities performed by 22 participants and categorized into three types: jogging, stretching, and strength exercises. The jogging category includes five labels: normal, rotating arms, skipping, sidesteps, and butt-kicks. The stretching category also comprises five labels: triceps, lunging, shoulders, hamstrings, and lumbar rotation. Lastly, the strength category encompasses eight labels: push-ups, push-ups (complex), sit-ups, sit-ups (complex), burpees, lunges, lunges (complex), and bench dips. In addition, the dataset includes a NULL label that represents unlabeled or undefined periods where no activity was explicitly carried out. It serves as a placeholder for missing annotations and can be useful for filtering, handling uncertainty, or distinguishing between known and unknown states. As a result,  participants engaged in 18 different outdoor sports activities. Unlike previous egocentric datasets, WEAR features an inertial sensor placement aligned with recent real-world application trends. It is the first dataset collected across multiple outdoor locations, each presenting varying surface conditions without revealing cues about the performed activities. Moreover, in FL in HAR applications, the selected dataset must mirror real-world conditions, as it should accommodate a large number of clients and a diverse array of classes and subjects. Herewith, based on the recent work \cite{geissler2024beyond}, the WEAR dataset remains the best option for fitness tracking, since other datasets cannot fulfill the needs due to having fewer participants, classes, and labeled data.

\vspace{-0.4cm}
\subsection{Hyperparameters and Framework}

This section provides a short summary of the most important parameters and framework used in our experiment.

\textbf{Client setting:} In the dataset, subjects 1 and 19, and also subjects 15 and 20, are from the same participants. These two participants were re-recorded in a different season and environment, resulting in 24 subjects in total. These two additional subjects are treated as separate devices for the same individuals to reflect real-world FitTech scenarios, thus setting the number of clients to 24.

\textbf{Data splitting:} In the WEAR dataset, participants perform the same activity several times in different time periods randomly, to represent realistic scenarios. This means that the activity patterns of people can change over time. For instance, while a person jogs fast at the beginning of exercise, he/she can feel tired later, which causes body movements to be slower. Furthermore, considering factors such as weather conditions and changes on the ground, it cannot be expected that a person will make the same body movements. Therefore, the method of splitting local data into training and testing requires special scrutiny. For this reason, we select the first 20 \% of each label in the time series to add to the testing data, and the remaining 80 \% is left for training. In this way, for example,  when a person jogs both at the start and later in the session, the test set will include both jogging bouts, ensuring temporal variation within each class.

\textbf{Window size:} After conducting multiple experiments, we determined that a window size of 100 is the optimal option, corresponding to a duration of 2 seconds in the time series sampled at 50 Hz.

\textbf{Batch size:} Despite the impact of other hyperparameters differing based on the used model and the dataset, \cite{kundroo2024demystifying} reported that a batch size of 32 emerged as the optimal choice for image data. Nevertheless, we set it to 32, since the hyperparameter recommendation for time series remains underexplored, and this value is used in the literature as well \cite{tu2021feddl}.

\textbf{Optimizer:} We employed the Adam (Adaptive Moment Estimation) optimizer because it has been shown to converge quickly and efficiently for complex models and large datasets. By adaptively tuning the learning rate of each parameter and using momentum to smooth the optimization, it requires less hyperparameter tuning, and these were found to be suitable for Federated Learning contexts \cite{li2021meta, presotto2022fedclar}.

\textbf{Learning rate:} Set to 0.001, following the literature \cite{xiao2021federated}.

\textbf{Communication round:} Based on the existing work in Federated Learning in HAR \cite{ek2022evaluation}, we have set the communication round to 100, as it is also a standard setting for other domains in FL in general.

\textbf{Local epoch:} Set to 1, following the literature \cite{cheng2023protohar, presotto2022fedclar}. 

\textbf{Model Aggregation:} We used FedAvg \cite{mcmahan2017communication} strategy since it is commonly employed as the main aggregation algorithm or a baseline in Federated Learning in HAR applications \cite{ek2022evaluation, li2021meta, cheng2023protohar, zhou2022you, chai2024profile, ouyang2021clusterfl, bu2024learn, tu2021feddl, presotto2022fedclar, xiao2021federated}.

\textbf{Early stopping:} Following the method in \cite{prechelt2002early}, we implemented early stopping based on the stability of the validation F1-score over a sliding window. The hyperparameters of this method are set based on our several experiments to determine the optimum values for this paper's case study. As a result of experimentation, the patience value is set to 5, and the threshold is set to 0.01 for the stopping criterion.

\textbf{Framework:} We use Flower (\url{https://flower.ai/}, \cite{beutel2020flower}), which is a widely adopted and openly available framework suitable for both industrial and academic research in Federated Learning. It is designed to simplify the implementation of Federated Learning by managing communication, orchestration, and model aggregation. Flower's support for large-scale experiments, including millions of simulated clients, demonstrates its remarkable scalability for real-world federated deployments. In addition, it offers broad compatibility with existing and emerging machine learning frameworks, diverse operating systems, and a wide range of hardware platforms, including servers and mobile devices. This high level of interoperability makes it a highly versatile tool for a variety of research applications.

\section{Evaluation Results}

The evaluation of this work compares FedFitTech with its client-side early stopping case study, considering both communication loads and global model performance on local data.

\begin{figure}[h]
    \centering
    \includegraphics[width=\linewidth]{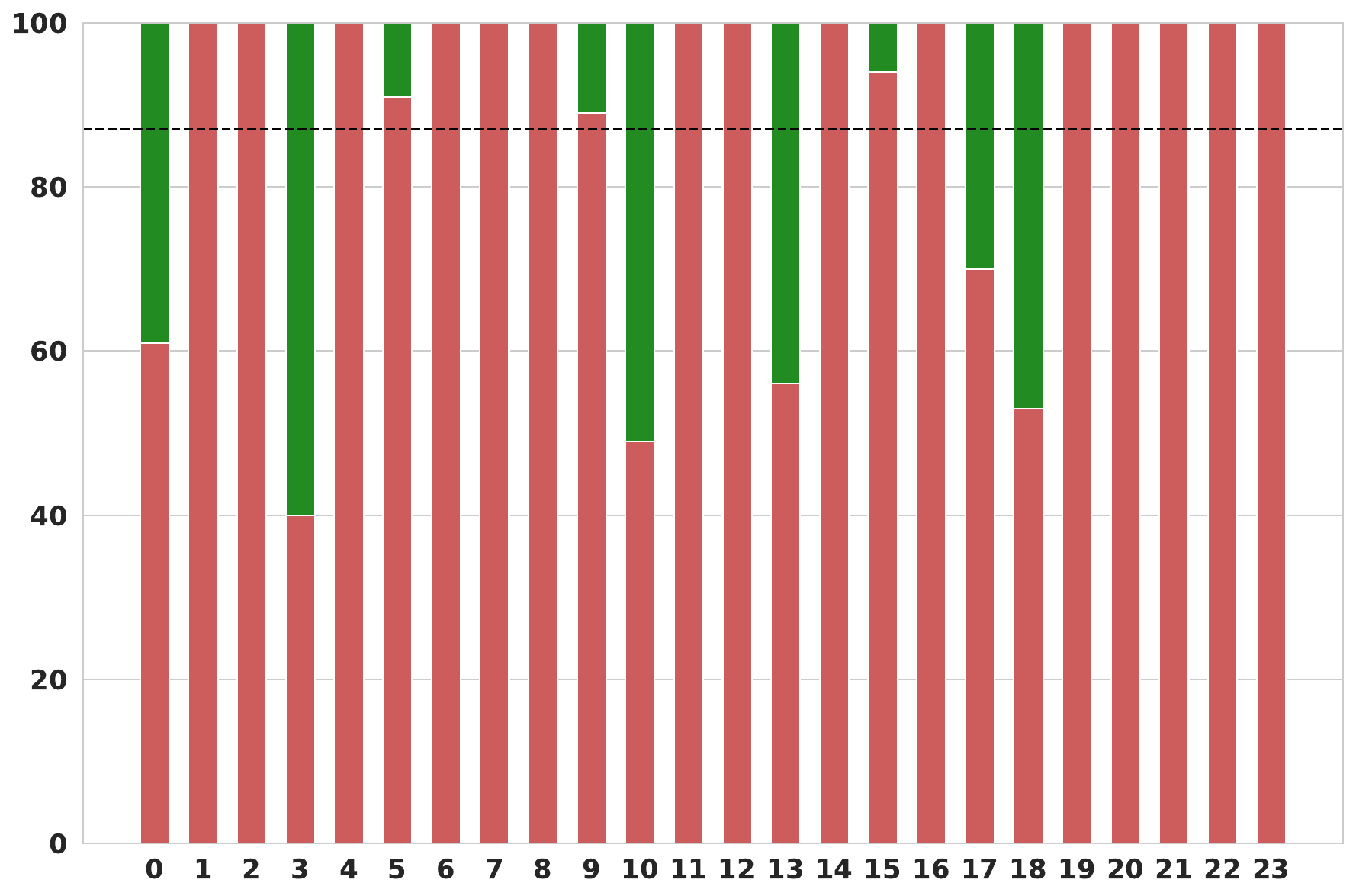}
    \caption{Amount of training rounds each client attended in the case study. The Y-axis shows the communication round, and the X-axis shows the client IDs. Red bars depict the communication cost, while green bars show the saved communication rounds, and the dashed black line is their mean.}
    \label{fig:GR_vs_clients_bar_chart}
\end{figure}

\subsection{Communication Cost}

Fig.~\ref{fig:GR_vs_clients_bar_chart} illustrates that 9 out of 24 clients stopped early from training the global model, highlighting that prolonged participation does not guarantee continued benefit. The earliest stop was observed at the 40th communication round (see client ID 3 in Fig.~\ref{fig:GR_vs_clients_bar_chart}), and some clients decided to stop early, even though they had participated in training the global model in more than 80 communication rounds.

\begin{figure}[h]
    \centering
    \includegraphics[width=\linewidth]{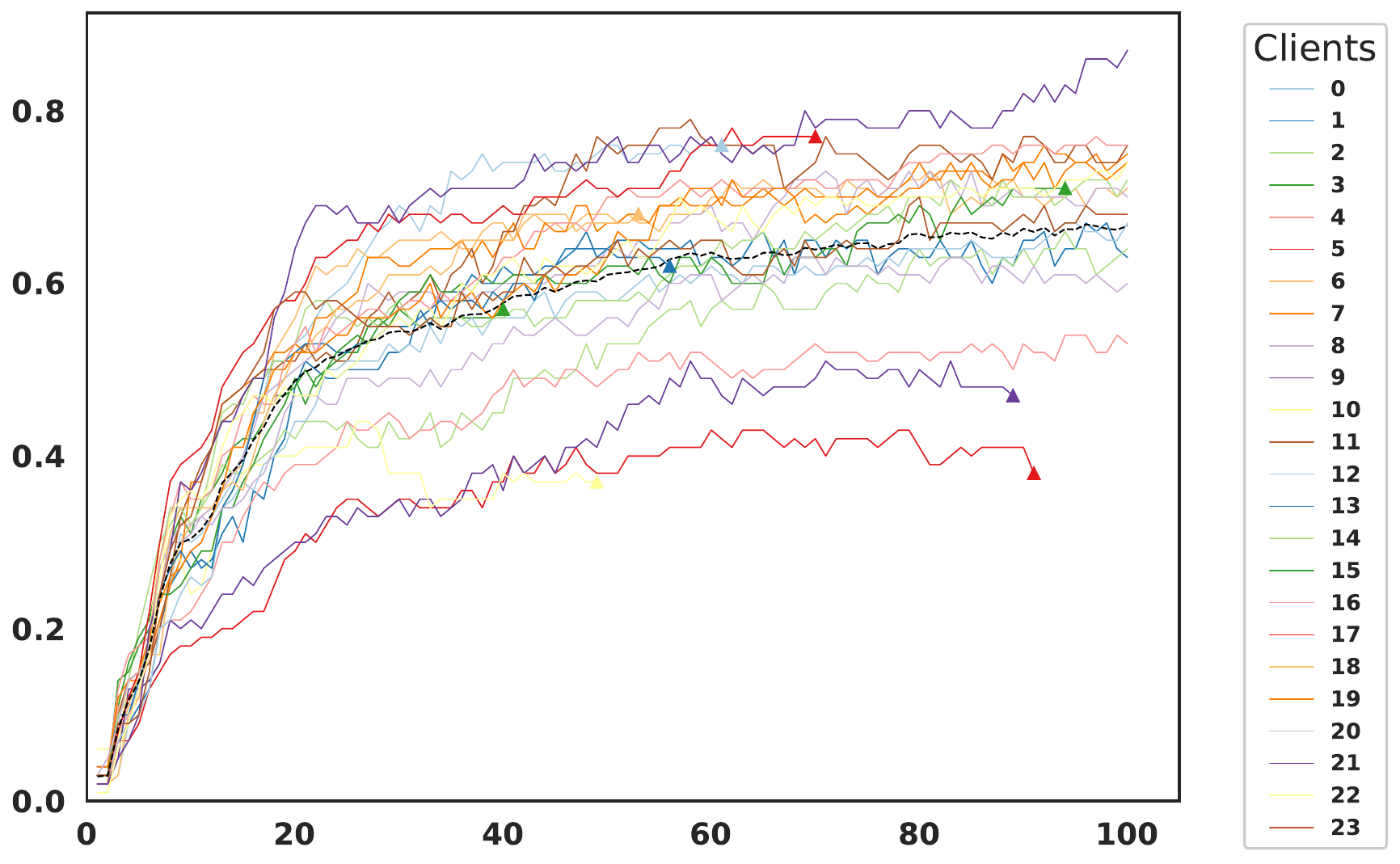}
    \caption{Communication rounds (X-axis) versus F1-scores (Y-axis). The dashed black line shows the mean values of all local performance, and the triangle markers depict rounds of early stopped clients.}
    \label{fig:F1-scores_convergence_line_graph}
\end{figure}

Moreover, Fig.~\ref{fig:F1-scores_convergence_line_graph} shows the performance of the F1-score of clients who are dropped early (see triangle markers within the plot). Some clients were unable to increase their local F1-score even after 30 rounds of training. Also, as can be seen from the black dashed line, the overall global model F1-score continues to increase, despite some clients being dropped.

\begin{figure}[h]
    \centering
    \includegraphics[width=\linewidth]{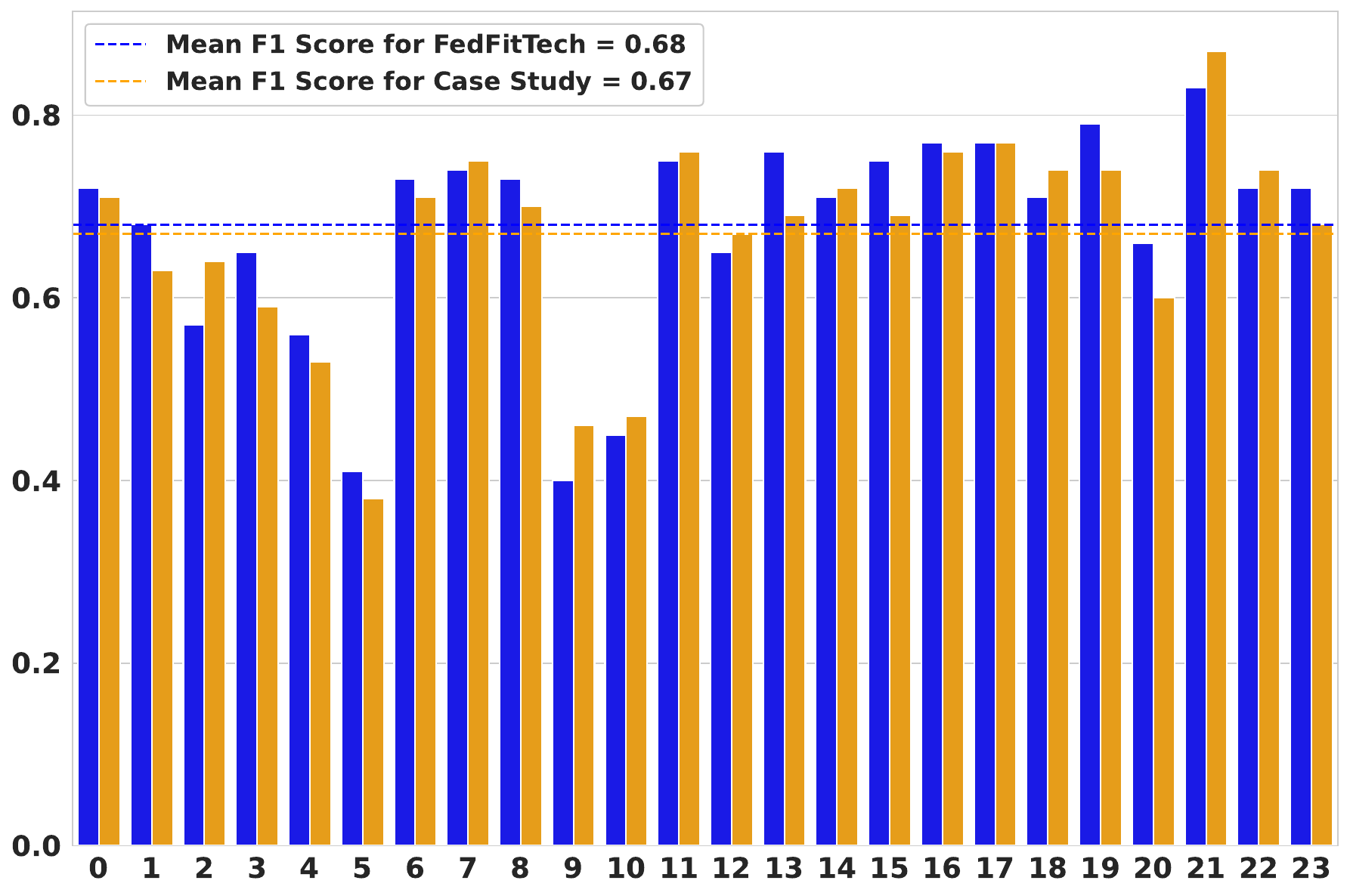}
    \caption{The different F1-scores (Y-axis) over all clients (X-axis) in the WEAR dataset, for the FedFitTech baseline and the early stopping case study experiment.}
    \label{fig:F1-scores_bar_chart}
\end{figure}

\subsection{Model Performance}
The client-based F1-scores are depicted in Fig.~\ref{fig:F1-scores_bar_chart} for both the baseline and our case study of early stopping. The FedFitTech baseline resulted in a 68\% mean F1-score across all clients, with the early stopping case study performing closely with a mean F1-score of 67\%. Additionally, the result presents that some of the clients even have better F1-scores in the system in the case study, such as the clients represented with these ID numbers: 2, 7, 9, 10, 11, 12, 14, 17, 19, 21, and 22, with 11 out of 24 clients obtaining a better performance compared to the FedFitTech baseline. In conclusion, the overall recognition performance remains almost unchanged, while early stopping reduces communication costs and overall computational load.

\begin{figure}[h]
    \centering
    \includegraphics[width=\linewidth]{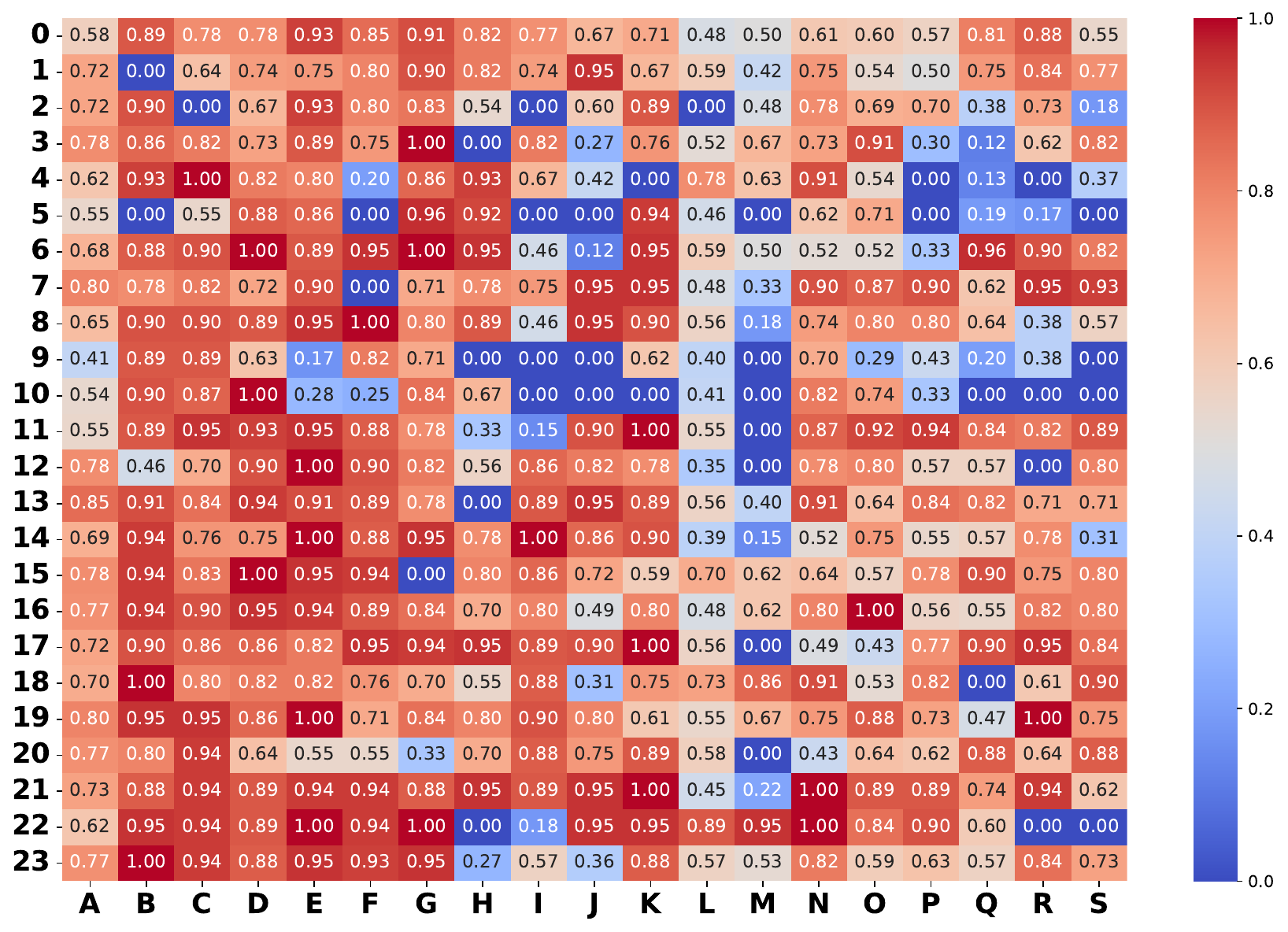}
    \caption{The client-label F1-scores of FedFitTech, with on the Y-axis the client IDs and on the X-axis the activity labels (as described in section 4.2).}
    \label{fig:confusion_matrix_FedFitTech}
\end{figure}

\begin{figure}[h]
    \centering
    \includegraphics[width=\linewidth]{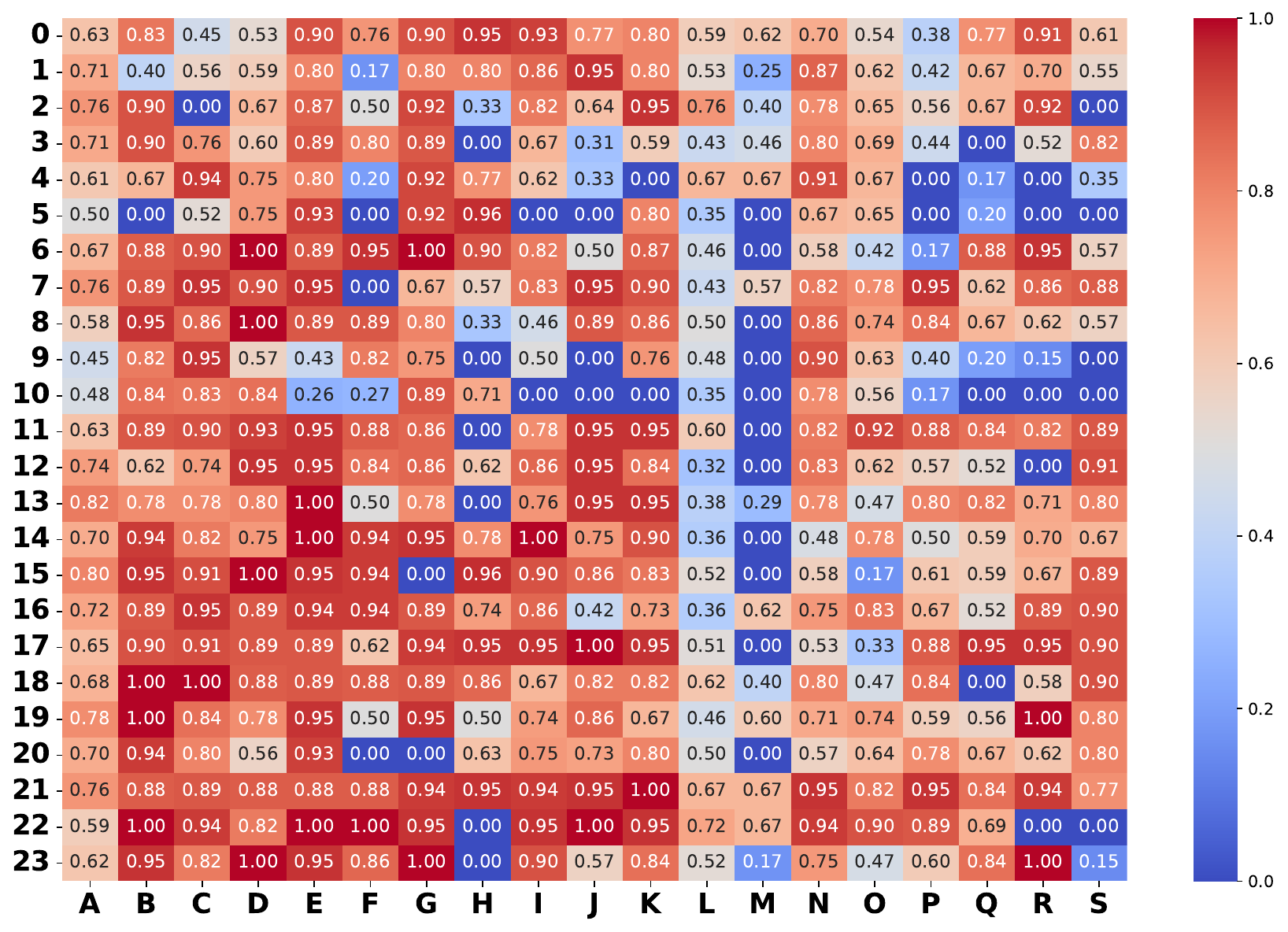}
    \caption{The client-label F1-scores for this paper's case study of early stopping. When comparing with Fig. \ref{fig:confusion_matrix_FedFitTech}, it can be seen that overall classification performance remains similar, with certain activities (such as L - push-ups and M - complex push-ups) remaining challenging across users. 
    } 
    \label{fig:confusion_matrix_case_study}
\end{figure}

Besides, Fig.~\ref{fig:confusion_matrix_FedFitTech} and Fig.~\ref{fig:confusion_matrix_case_study} depict the label-based F1 scores of the FedFitTech baseline and the case study, respectively. Actual label names in these figures are as follows: 'A': 'NULL', 'B': 'jogging', 'C': 'jogging (rotating arms)', 'D': 'jogging (skipping)',  'E': 'jogging (sidesteps)', 'F': 'jogging (butt-kicks)', 'G': 'stretching (triceps)', 'H': 'stretching (lunging)', 'I': 'stretching (shoulders)', 'J': 'stretching (hamstrings)', 'K': 'stretching (lumbar rotation)', 'L': 'push-ups', 'M': 'push-ups (complex)', 'N': 'sit-ups', 'O': 'sit-ups (complex)', 'P': 'burpees', 'Q': 'lunges', 'R': 'lunges (complex)', and 'S': 'bench-dips'.

Despite early stopping decreasing the overall mean F1-score slightly by 1\% (see Fig.~\ref{fig:F1-scores_bar_chart}), some of the clients reached a remarkable performance increase for some of the labels. For instance, the push-up fitness activity (see label L in the confusion matrices) had an F1-score for Client 2 in FedFitTech of 0\%, while it is 76\% in the case study. Client 18, which was the one who stopped training the first (see Fig.~\ref{fig:GR_vs_clients_bar_chart}), achieved a 47\% reduction in communication costs, and its overall F1-score increased from 71\% to 74\%. Also, tracking some of the activities (e.g., 'stretching (lunging)': H) increased from 55\% to 86\%.

\subsection{Discussion}
Fig.~\ref{fig:GR_vs_clients_bar_chart} and~\ref{fig:F1-scores_convergence_line_graph} show that this client attrition, while seemingly modest in this small-scale experiment, represents a significant concern in real-world Federated Learning deployments involving a large number of clients, as 37.5\% of clients decreased their communication loads. Besides, these findings suggest that client-specific factors, such as data heterogeneity, resource constraints, or local model convergence, may impact the stability of participation within the Federated Learning framework. However, as shown in Fig.~\ref{fig:F1-scores_convergence_line_graph}, the global model continues to improve despite some clients no longer participating in training. Also, Fig.~\ref{fig:F1-scores_bar_chart}  demonstrates that in the case study, 45.8\% of the clients increased their overall mean F1-scores, while communication costs decreased by 13\% (see Fig.~\ref{fig:GR_vs_clients_bar_chart}).


\section{Conclusions and Future Work}
Wearable devices provide rich data that enable in-depth insights into user fitness activities within the domain of Fitness Technology (FitTech). However, manual data annotation in FitTech is both time‑consuming and prone to error, creating a significant bottleneck. Users can offload the tedious task of annotating activities while benefiting from diverse and high quality labels. Although Centralized Learning approaches can harness all this annotated data, they often struggle with latency, limited scalability, and heightened privacy risks.

We argue in this paper that Federated Learning aligns well with the characteristics of FitTech, as clients have largely similar sensors, similar processing power, and mostly common activity classes, while requiring relatively infrequent communication between a global model and the clients. This work thus introduces FedFitTech, designed to explore the FitTech domain within Federated Learning, built upon the user-friendly, widely adopted, and open-source Flower framework. Moreover, we shared a case study as an example usage of this baseline, which involves a client-side early stopping strategy. The case study is compared with the FedFitTech baseline, considering communication loads and model performance metrics. The findings show that our case study mitigates the communication burden with a negligible mean F1-score drop by balancing model generalization.

The FedFitTech baseline provides various promising directions for future research in Federated Learning for the FitTech domain. For instance, exploring transfer learning implementations can accelerate convergence and enable clients to benefit from pre-existing patterns from the outset. Additionally, integrating differential privacy techniques is crucial for improving data privacy and security, addressing a fundamental concern in FitTech. Furthermore, since the utilized WEAR dataset in FedFitTech consists of both inertial and ego-perspective camera data, investigating multimodal mechanisms has significant potential for handling diverse sensor data and mitigating data heterogeneity. 

Moreover, exploring efficient local model clustering methods and then sharing cluster-specific global models with clients could reduce performance differences across clients and increase overall recognition performance by facilitating collaboration among users with similar behavioral patterns (e.g., activity routines, age, weight, or gender). In conclusion, we hope that our FedFitTech baseline will leverage the publicly available and widely used Flower framework, opening up a new research domain and development opportunities in FitTech.

The FedFitTech baseline is available as an open source repository at: \url{https://github.com/shreyaskorde16/FedFitTech}{}

\textbf{Acknowledgments}
This work was funded by the Deutsche Forschungsgemeinschaft (DFG, German Research Foundation) – Project-IDs 506589320 and 520256321.


\end{document}